\documentclass[10pt,twocolumn,letterpaper]{article}
\usepackage{amssymb}
\usepackage{wacv}
\usepackage{pbox}
\usepackage{times}
\usepackage{epsfig}
\usepackage{graphicx}
\usepackage{amsmath}
\usepackage{amssymb}
\usepackage{amsfonts}
\usepackage{ulem}
\usepackage{multicol}
\usepackage{multirow}
\usepackage{graphics}
\usepackage{pifont}
\usepackage{gensymb}
\usepackage{authblk}

\def\wacvPaperID{1057} 

\wacvfinalcopy 

\ifwacvfinal
\def\assignedStartPage{1} 
\fi

\usepackage{xcolor}
\usepackage{makecell}
\newcommand{\xmark}{\ding{55}}%

\ifwacvfinal
\usepackage[breaklinks=true,bookmarks=false]{hyperref}
\else
\usepackage[pagebackref=true,breaklinks=true,colorlinks,bookmarks=false]{hyperref}
\fi

\ifwacvfinal
\else
\fi

\begin{document}

\title{\textit{DUDE}: Deep Unsigned Distance Embeddings for Hi-Fidelity Representation of Complex 3D Surfaces

}

\ifwacvfinal
\author[1]{Rahul Venkatesh}
\author[2]{Sarthak Sharma}
\author[3]{Aurobrata Ghosh}
\author[4]{Laszlo Jeni}
\author[5]{Maneesh Singh}
\affil[1,4]{Carnegie Mellon University, Pittsburgh, PA, USA}
\affil[2,3]{Verisk Analytics, Hyderabad, Telangana, India}
\affil[5]{Verisk Analytics, Jersey City, NJ, USA}
\fi

\maketitle
\begin{figure*}
    \centering
    \includegraphics[scale=0.21]{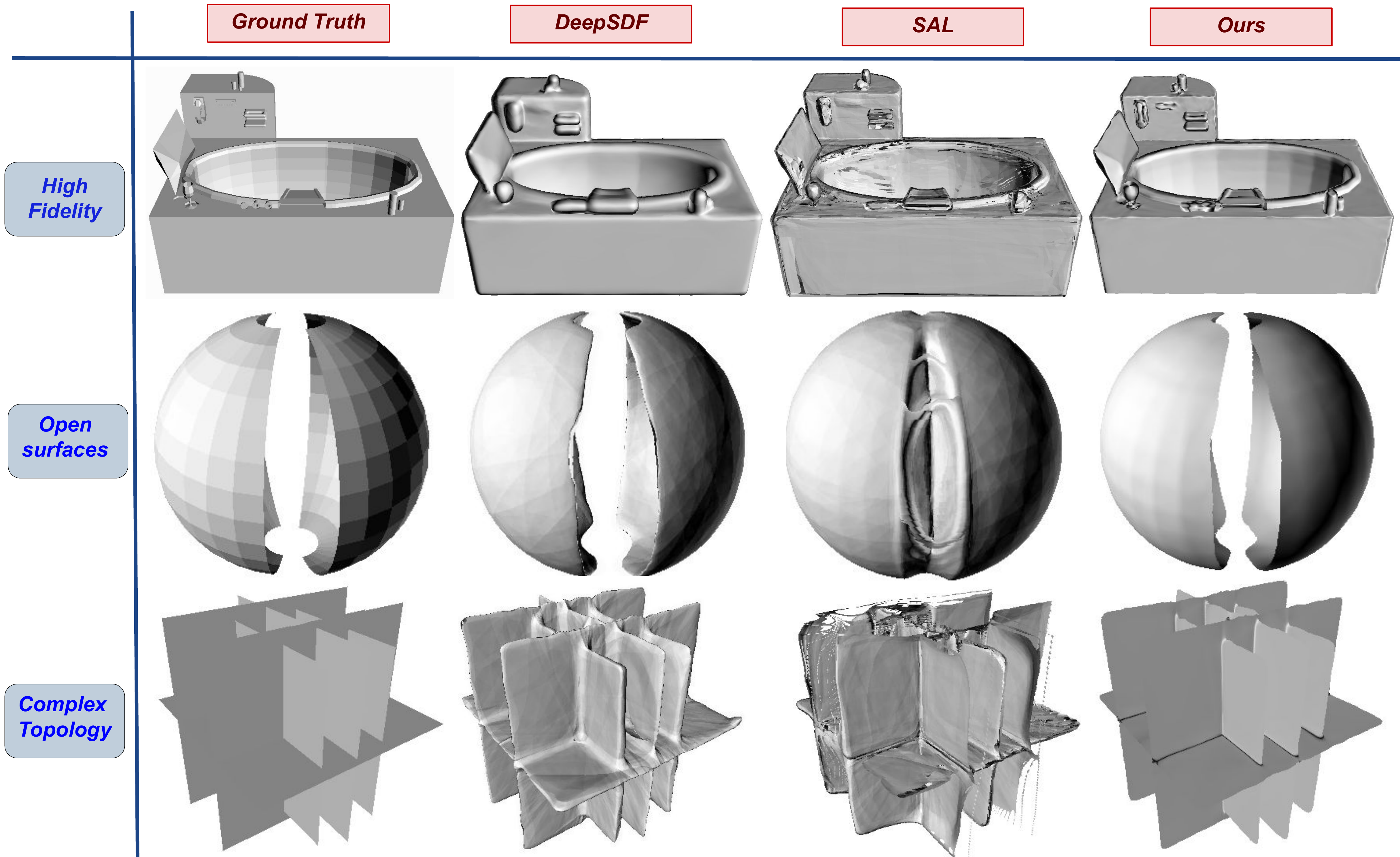}
    \caption{Implicit 3D Shape Representations learnt on three exemplar shapes each highlighting an important property of the representation. Note that we evaluate our method with DeepSDF~\cite{deepsdf} and SAL~\cite{sal-2020}. While DeepSDF learns over-smoothed representations \textit{\textbf{(Top Row)}} owing to the intermediate watertighting step, SAL learns to close out regions which are intended to be open \textit{\textbf{(Middle Row)}}.   Additionally, both DeepSDF and SAL have poor surface normal estimates on Open Surfaces (Middle Row) and Complex Topology \textit{\textbf{(Bottom Row)}}. In contrast, our method learns high fidelity shape representations while at the same time generating visually pleasing renderings on complex shapes owing to the high quality normals estimated by a separately learnt normal vector field (\textit{nVF}). Refer Section.~\ref{sec:experiments} for more details.}
    \label{fig:concept}
\end{figure*}

\begin{abstract}

    High fidelity representation of shapes with arbitrary topology is an important problem for a variety of vision and graphics applications. Owing to their limited resolution, classical discrete shape representations using point clouds, voxels and meshes produce low quality results when used in these applications. Several implicit 3D shape representation approaches using deep neural networks have been proposed leading to significant improvements in both quality of representations as well as the impact on downstream applications. However, these methods can only be used to represent topologically closed shapes which greatly limits the class of shapes that they can represent. As a consequence, they also often require clean, watertight meshes for training. In this work, we propose DUDE - a Deep Unsigned Distance Embedding method which alleviates both of these shortcomings. DUDE is a disentangled shape representation that utilizes an unsigned distance field (\textit{uDF}) to represent proximity to a surface, and a normal vector field (\textit{nVF}) to represent surface orientation. We show that a combination of these two (\textit{uDF+nVF}) can be used to learn high fidelity representations for arbitrary open/closed shapes. As opposed to prior work such as DeepSDF, our shape representations can be directly learnt from noisy triangle soups, and do not need watertight meshes. Additionally, we propose novel algorithms for extracting and rendering iso-surfaces from the learnt representations. We validate DUDE on benchmark 3D datasets and demonstrate that it produces significant improvements over the state of the art.
\end{abstract}



\section{Introduction}
\label{sec:intro}
High fidelity representation of potentially open 3D surfaces with complex topologies is important for reconstruction of 3D structure from images, point clouds and other raw sensory data, fusion of representations from multiple sources and for rendering of such surfaces for a lot of applications in vision, graphics and the animation industry ~\cite{survey3D}. 

Classical shape representations using triangle or quad meshes pose challenges in reconstructing surfaces with arbitrary topology ~\cite{pan2019deep}. In addition, the resolution of such reconstructions is limited by the predefined number of vertices in the network. Recent efforts seek to represent surfaces using implicit functions represented using deep neural networks ~\cite{deepsdf,occupancy_network, cvxnet-2020, chen2020bsp, sal-2020}. Deep neural networks can, in principle, represent shapes with arbitrary topology to any degree of accuracy (limited by network capacity).

However, current approaches are unable to model surfaces that are open or noisy input data containing holes in the surfaces. Approaches, such as~\cite{deepsdf}, learn the Signed Distance Function (SDF $\mathcal{F}$) as the implicit function from ($p_i$, $\mathcal{F}(p_i)$) samples where the SDF $\mathcal{F}(p_i)$ is positive (negative) for points $p_i$ inside (outside) the surface. This requires that the ground truth surface be watertight (closed). Since most 3D shape datasets~\cite{chang2015shapenet} do not have watertight shapes, preprocessing is needed to create watertight meshes ~\cite{occupancy_network} which is known to result in loss of surface fidelity~\cite{huang2018robust}. Other methods, e.g. ~\cite{sal-2020}, learn implicit surface representations directly from raw unoriented point clouds. However, such methods also make an assumption that the underlying surface represented by the point cloud is closed, leading to learnt representations necessarily describing closed shapes. As pointed out in~\cite{sal++}, even in cases where the raw input point cloud is scanned from an open surface, learnt representation tends to incorrectly close the surface. 
Since current approaches assume that the 3D shaped to be modeled are closed,
they suffer from a loss of fidelity when modeling open shapes or learning from noisy meshes. 

To address these shortcomings, this work proposes a novel approach satisfying the following properties. (1) \textbf{Representation Power:} Capability of accurately modeling both open and closed shapes, with arbitrary topology. (2) \textbf{Learning from Noisy Meshes:} As noted in~\cite{sal-2020}, most raw scan data is stored in the form of raw triangle soups, and not watertight meshes that are used in~\cite{deepsdf}. The proposed approach should allow learning directly from such inputs.   


\noindent Our approach comprises of the following components: \\
\noindent
\textbf{Unsigned Distance Functions.} Since signed distance functions are only defined for closed shapes (shapes that partition the 3D space into interior and exterior regions), we propose to model unsigned distance function (\textit{uDF}) instead, which can be unambiguously defined for both open and closed shapes~\cite{mullen2010signing}. 

\noindent
\textbf{Surface Normals.} Many applications also require the easy availability of surface normals. Examples include rendering using ray tracing and optimizing for downstream tasks like shape retrieval ~\cite{liu2020dist}. However, this leads to the following challenge: while \textit{sDF}s are differentiable everywhere, allowing for the extraction of surface normals from the learnt representation in a differentiable manner~\cite{liu2020dist}, uDFs are not differentiable at the surface (differentiable everywhere else). Consequently, the extraction of surface normals using differential operators is not stable.   

Firstly, since the surface normal can still be defined as some (unknown) function of the uDF, we can also model it reliably using a DNN irrespective of the non-differentiability of the uDF on the surface. Hence, we propose to learn a normal vector field (\textit{nVF}) using a DNN which also makes available the normal to the learnt surface. Thus, we have decomposed the implicit representation of a shape into two parts: (1) a \textit{uDF}, and, (2) an \textit{nVF}. We demonstrate that these together are capable of accurately representing any arbitrary shape with complex topology, irrespective of whether it is open/closed (in contrast to existing implicit shape representations~\cite{deepsdf, occupancy_network, cvxnet-2020, sal-2020, sal++, chen2020bsp}). Secondly, while the \textit{nVF} produces a continuous normal vector field modeling normals to an unoriented surface, we seek to learn it directly from noisy triangle soups with oriented normals. This causes the \textit{modulo $180\degree$ problem} when computing the error between \textit{nVF} and the normals in the data. We therefore propose a robust loss function to train the \textit{nVF}. 

\noindent
\textbf{Isosurface Extraction.} Once the implicit surface representation is learnt, it is often required to extract the explicit representation of the surface ~\cite{akkouche2001adaptive}. To support this functionality, we design an efficient algorithm to perform multi-resolution iso-surface extraction from \textit{uDFs}.

\noindent
\textbf{Rendering.} We also propose a novel sphere tracing algorithm that utilizes the learnt \textit{nVF} to enable more accurate ray-scene intersection. Note that \textit{sDF}s allow for precise computation of ray-scene intersections using a bisection search close to the surface. However, such an algorithm cannot be applied to \textit{uDFs} owing to the fact that function doesn't change sign on crossing the surface. The proposed algorithm choosing a (conservative) threshold while sphere tracing \textit{uDFs} to get close to the surface and then utilizes the learnt \textit{nVF} close to the surface for accurate ray-scene intersections. 

The above algorithms are described in Section.~\ref{sec:spheretracing}. 

To summarize, our \textbf{contributions} are: (1) DUDE - A deep, unsigned distance embedding approach capable of high fidelity representation of both open and closed shapes with complex topology, from raw triangle soups. (2) An efficient multi-resolution iso-surface extraction technique to extract meshes from \textit{uDFs}. (3) A novel \textit{uDF} sphere tracing algorithm that leverages a normal vector field defined over the surface. 


\begin{table}[h!]
\resizebox{\columnwidth}{!}{
\begin{tabular}{c|c|c|c|c}
\hline
    \multirow{2}{*}{Method}& \multirow{2}{*}{\makecell{ Learning \\ from\\ Triangle Soups}}&\multicolumn{3}{c}{Representation Power} \\
    \cline{3-5}
     &  & \makecell{Open\\Shapes} & \makecell{Complex \\ Topology} & \makecell{High \\ Fidelity} \\
    \hline
     DeepSDF~\cite{deepsdf}& \xmark& \xmark & \checkmark & \checkmark  \\
     SAL~\cite{sal-2020} & \checkmark & \xmark & \checkmark & \checkmark  \\
     \hline
     DUDE (Ours) & \checkmark & \checkmark & \checkmark & \checkmark   \\
\hline
\end{tabular}
}
\vspace{0.5mm}
\caption{Comparison between DUDE and the closely related art.}
\vspace*{-2mm}
\label{table:rep power}
\end{table}

It's illustrative to compare DUDE with the closely related art outlined in Table.~\ref{table:rep power} which shows that while DeepSDF~\cite{deepsdf} can't work with raw triangle soups, both DeepSDF and SAL~\cite{sal-2020} can't represent open shapes while the proposed approach can learn high fidelity representations of open (and closed) shapes with complex topologies, directly from raw triangle soups. We illustrate these differences qualitatively in Fig.~\ref{fig:concept}. The remainder of the paper is organized in the following way. We formalize the approach in Section.~\ref{sec:approach} after first introducing the related work in this field in Section~\ref{sec:related_wk}. Later, we describe algorithms for iso-surface extraction in Section.~\ref{sec:MISE}, and finally, provide an empirical justification for our representation in Section.~\ref{sec:experiments}.  


\section{Related Work}
\label{sec:related_wk}

\noindent
\textbf{Explicit Shape Representations.} 
The traditional methods of 3D shape representation are voxels and point clouds. Voxels provide a direct extension of pixels to 3D, allowing easier extension of image processing techniques for shape analysis. Several initial shape representation works are built upon this idea \cite{voxnet-2015,choy2016,pix3d}. 
Drawbacks of the voxel-representation are higher computational/memory requirements and limited output resolution. Polygonal mesh representations addresses some of these issues
\cite{atlasnet2018,wang2018pixel2mesh,meshrcnn-2019}, although they are still limited by number of vertices and topology. 
Point clouds provide a more compact and sparser encoding of surface geometry. Points on the object surface can either be directly obtained using depth sensors \cite{kinectfusion-2011} or estimated from multi-view images 
\cite{LearningMultiview-2020}. 
Recently several deep learning methods have been proposed which directly work on points as inputs like \cite{achlioptas2017latent_pc,Guo2020DeepLF}. Though more efficient than voxels, point clouds provide limited output resolution and do not encode a topology.


\noindent
\textbf{Implicit Shape Representations.}
Deep neural networks (DNN) generalise implicit functions in two ways: classification and regression. A DNN is trained either to (1) classify a point as inside/outside a shape~\cite{occupancy_network}  (delineated by the modeled surface):
$\mathcal{N}_c:\mathbb{R}^3\rightarrow\{0,1\}$; or to (2) regress the signed
distance of the point to the surface~\cite{deepsdf}
$\mathcal{N}_r:\mathbb{R}^3\rightarrow s\in\mathbb{R}$, the sign of 
$s$ indicating inside or outside.

Hybrid explicit/implicit representations are proposed \cite{chen2020bsp,cvxnet-2020}, where the implicit function is a union of inside/outside classifier hyper-planes. BSP-Net \cite{chen2020bsp} proposes
a binary space partitioning network to model a convex decomposition of the 3D shape, the union of which defines a watertight separation of the inside/outside of the shape.
CvxNet \cite{cvxnet-2020}, also proposes a convex decomposition using 
hyper-planes but with a double representation of a complex primitive.

Above methods only represent watertight shapes. Hence, the training data is also expected to satisfy this condition. However, not all sources of 3D shapes are watertight and a water-tightening pre-processing step often results in non-trivial
loss of fine details \cite{occupancy_network, deepsdf}. ShapeNet \cite{chang2015shapenet}, for example, comprises of CAD models
that are neither watertight nor composed of non-intersecting meshes.
Other sources include point clouds and triangle soups. Sign agnostic learning (SAL) \cite{sal-2020} addresses this limitation by training an implicit representation on an unsigned distance loss. However, this results in incorrect surface reconstruction for open shapes (Fig.~\ref{fig:concept}). In contrast, our method can represent open (or closed) complex shapes with high fidelity.

\begin{figure}
    \centering
    \includegraphics[scale=0.33]{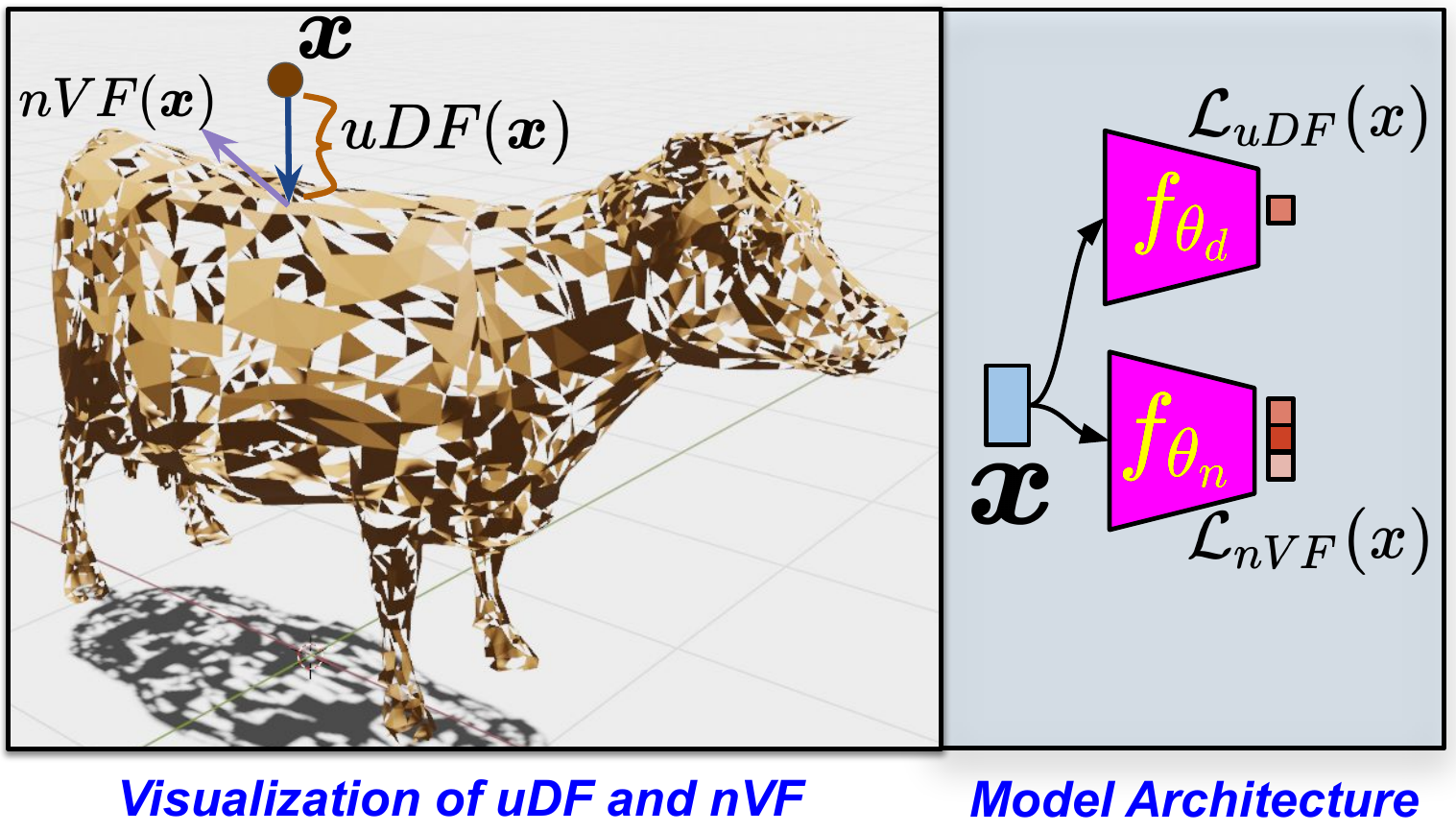}
    \caption{\textbf{\textit{Left:}} An illustration of what the \textit{uDF} and \textit{nVF} represent. \textit{\textbf{Right:}} Model architecture used for training.} 
    \label{fig:udf_nvf}
\end{figure}

\section{Approach/Method}
\label{sec:approach}

In this section, we describe the proposed disentangled shape representation, followed by a description of the training algorithm. 

\subsection{Shape Representation}
\label{sec:shape_rep}
We disentangle the implicit representation of a shape into  an unsigned distance field (\textit{uDF}) and a normal vector field (\textit{nVF}). As discussed in Section.~\ref{sec:intro}, this decomposition enables us to represent both open and closed shapes with arbitrary topology, which, to the best of our knowledge, is a significant improvement over existing methods. Similar to ~\cite{deepsdf} we aim to model these functions using feed-forward networks with non-linear activation functions. \\

We begin with a formal definition of a \textit{uDF}: it's a function that outputs the closest unsigned distance to the surface from any given point in 3D space. Note that this is in contrast to an \textit{sDF} which is meant to output negative distances inside the shape and positive distances outside. Several notable current approaches~\cite{deepsdf, sal-2020, sal++} use \textit{sDFs} for representing shapes, thus bound to the assumption that the underlying surface is watertight. 
To relax this assumption, we model a 3D shape using a \textit{uDF} which can represent both watertight and non-watertight shapes equally well.    
\begin{equation}
    uDF(\boldsymbol{x}) = d : x \in \mathbb{R}^3, d \in \mathbb{R}^+.
\end{equation}

\begin{figure}[b]
    \centering
    \includegraphics[width=\columnwidth]{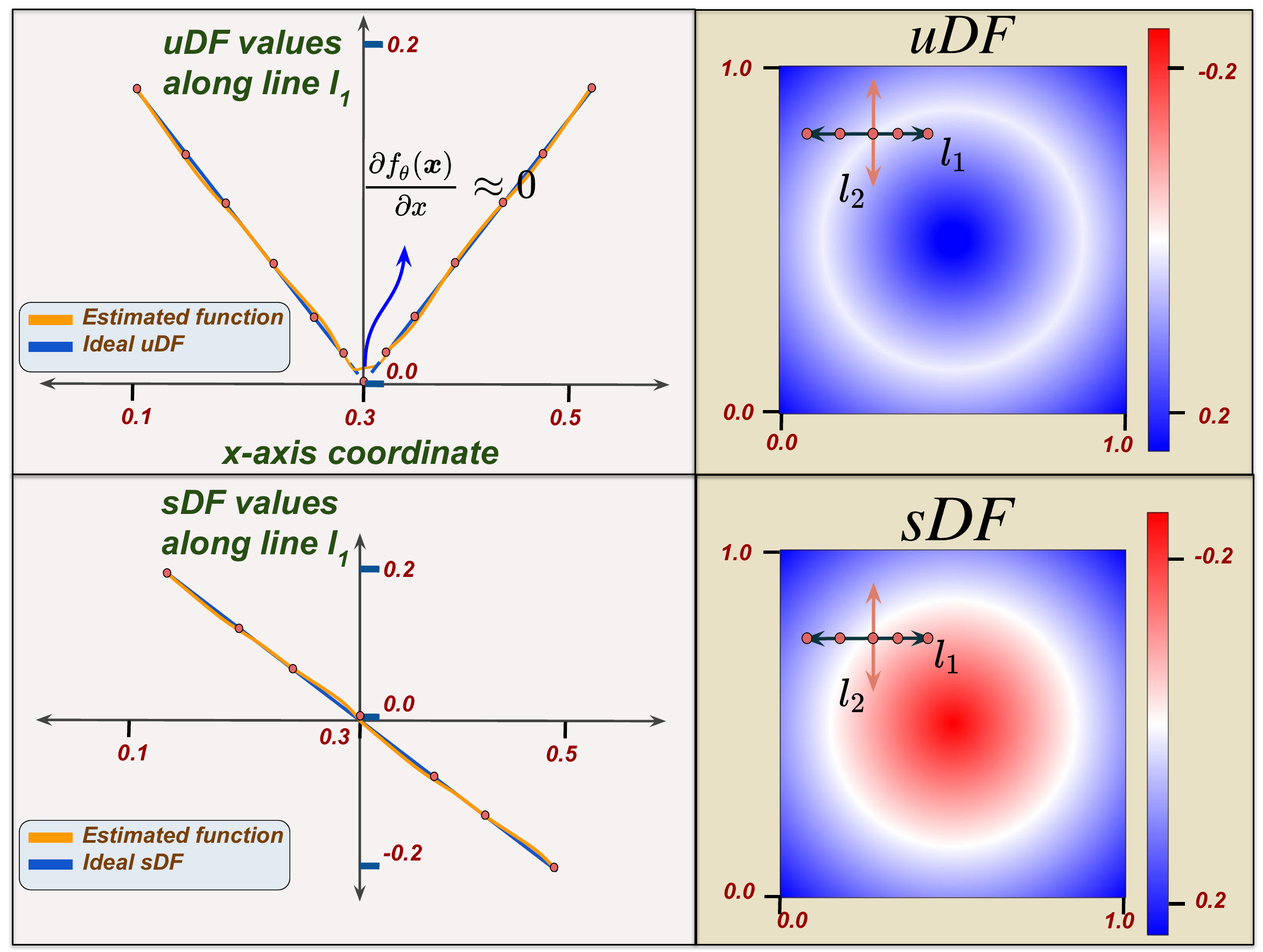}
    \caption{Graph representing non-differentiability of the \textit{uDF} on the surface of the shape. \textit{\textbf{Right:}} Visualization of the distance field with sampled points on line $l_1$. \textit{\textbf{Left:}} Graphs of the \textit{uDF} and \textit{sDF} values sampled along line $l_1$ for both the estimated and the ideal function.}
    \label{fig:udf_graph}
\end{figure}

As can be noted, trivially removing the sign of the \textit{sDF} leads to a \textit{uDF}. However, this sacrifices some important properties of the \textit{sDF} leading to a new set of challenges which we need to address. (1) The \textit{uDF} is non-differentiable at the surface (see Figure~\ref{fig:udf_graph}) implying it can't be reliably trained using points sampled from the surface. We overcome this by never actually sampling the training data points on the surface but slightly away from the surface; (2) Unlike \textit{sDFs}, we cannot reliably extract the surface normal from a \textit{uDF} due to its non-differentiability (See Figure~\ref{fig:udf_graph}). Since estimation of high quality surface normals is important for several downstream tasks~\cite{liu2020dist}, we propose to learn a normal vector field (\textit{nVF}), which, for any 3D location, $x$, represents the \textit{nVF} normal to the surface point closest to $x$. Formally,
\begin{equation}
\begin{gathered}
    nVF(\boldsymbol{x}) = \boldsymbol{v}: \boldsymbol{x} \in \mathbb{R}^3, \boldsymbol{v} \in \mathbb{R}^3, \\
    \boldsymbol{v} = n(\boldsymbol{\tilde{x}}): \boldsymbol{\tilde{x}} = \boldsymbol{x} + \boldsymbol{r_x}*uDF(\boldsymbol{x}) .
\end{gathered}
\end{equation}

\noindent where $\boldsymbol{r_x}$ is the unit vector from the point $\boldsymbol{x}$ to its closest point on the surface, i.e. $\boldsymbol{\tilde{x}}$. Additionally, $n(\boldsymbol{x})$ is the normal to the surface at the point $\boldsymbol{x}$. We model the \textit{uDF+nVF} pair using MLPs which we aim to train using a noisy triangle soup, a noisy representation of the underlying ground truth surface. 

Given a 3D shape represented by the noisy triangle soup, we construct training samples, $\mathcal{P}$, which contain a point, $\boldsymbol{x}$, the \textit{uDF} and the \textit{nVF} evaluated at $\boldsymbol{x}$
\begin{equation}
    \mathcal{P} = \{(\boldsymbol{x}, d, \boldsymbol{v}): d = uDF(\boldsymbol{x}), \boldsymbol{v}=nVF(\boldsymbol{x})\}.
\end{equation}

\noindent using the following procedure: we first densely sample a set of \{points, surface normal\} pairs from the triangle soup, by uniformly sampling points on each triangle face. Let's call this set of points $\mathcal{X}= \{(\boldsymbol{x_s}, \boldsymbol{v_s})\}$. Since each point is sampled from a triangle face, the normal to the triangle face provides the associated surface normal for that point. 

\begin{figure}
    \centering
    \includegraphics[scale=0.4]{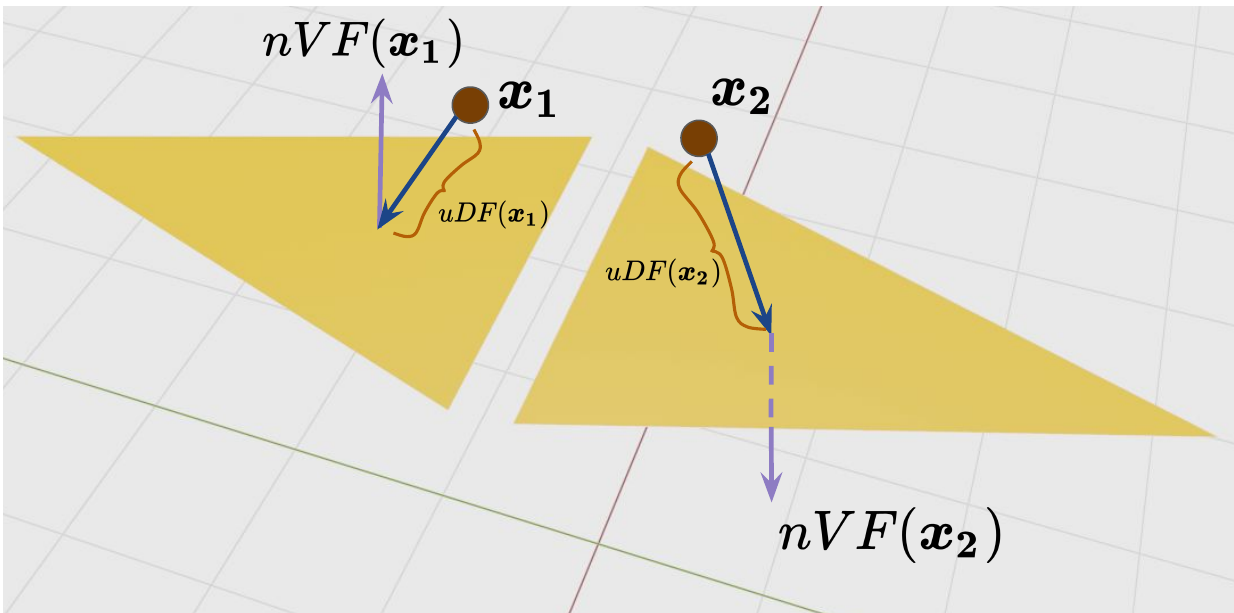}
    \caption{Motivation for design of $\mathcal{L}_{nVF}$ (See Section.~\ref{sec:shape_rep}). For a triangle soup, adjacent faces can have oppositely facing normals.}
    \label{fig:normals}
\end{figure}

Given this set $\mathcal{X}$, the set $\mathcal{P}$ is constructed by sampling points $\boldsymbol{x}$ in 3D space and finding the nearest corresponding point in $\mathcal{X}$ to construct the training sample $(\boldsymbol{x}, ||\boldsymbol{x_s}-\boldsymbol{x}||_2, \boldsymbol{v_s})$. The set $\mathcal{P}$ is used train the DNNs to approximate the \textit{uDF} and the \textit{nVF}. More concretely, we train the DNN $f_{\theta_d}$ to approximate \textit{uDF}, and DNN $f_{\theta_n}$ to approximate \textit{nVF}.

The loss function used to train $f_{\theta_d}$ is straightforward. We simply use the \textit{L2} loss between the estimated unsigned distance, $f_{\theta_d}(\boldsymbol{x})$ and the ground truth distance, $d = ||\boldsymbol{x_s}-\boldsymbol{x}||_2$. 
\begin{equation}
    \mathcal{L}_{uDF} = ||f_{\theta_d}(x) - d||_2.
\end{equation}


Before we describe how $f_{\theta_n}$ is trained, please note that \textit{uDF}s naturally correspond to unoriented surfaces (which are also logically necessitated by open surfaces). However, for most ray-casting applications this is not an issue as the direction of the first intersected surface can be chosen based on the direction of the ray. So, the ambiguity of $n$ or $-n$ can be handled. This implies a \textit{modulo 180\degree} representation in the DNN suffices. However, such a representation needs to be learnt from a noisy triangle soup with oriented surface normals with possible directional incoherence (in the \textit{modulo 180\degree} sense) between adjacent triangles. To allow for this, we optimize the minimum of the two possible losses, computed from each $n$ or $-n$. More concretely,   
\begin{equation}
    \begin{gathered}
      \mathcal{L}_{nVF}^{(1)} = ||f_{\theta_n}(x) - v_s||_2,\\
      \mathcal{L}_{nVF}^{(2)} = ||f_{\theta_n}(x) - (-v_s)||_2,\\
      \mathcal{L}_{nVF} = min(\mathcal{L}_{nVF}^{(1)},\mathcal{L}_{nVF}^{(2)}).  
    \end{gathered}
\end{equation}

This allows for the network to learn surface normals \textit{modulo 180\degree}. The incoherence in the noisy triangle soup is handled by the continuity property of the DNNs and, practically, coherent normal fields are learnt as verified in our experiments. 
Thus, after training, the zero-level set of $f_{\theta_d}$, which approximates the \textit{uDF}, represents points on the surface, while $f_{\theta_n}$, approximating \textit{nVF}, represents the surface normals of the corresponding points on this level set. 



\section{Iso-surface Extraction and Rendering}
\label{sec:MISE}
In this section, we describe a technique to extract an iso-surface out of the learnt representation. This process is important for many downstream vision applications such as shape analysis~\cite{laga20183d} and graphics applications such as creation and rendering of novel scenes using ray tracing under changed illumination, texture or camera viewpoints ~\cite{purcell2005ray}. 

\subsection{Sphere Tracing uDFs.} 
\label{sec:spheretracing}Sphere tracing~\cite{hart1996sphere} is a standard technique to render images from a distance field that represents the shape. To create an image, rays are cast from the focal point of the camera, and their intersection with the scene is computed using sphere tracing. Roughly speaking, irradiance/ radiance computations are performed at the point of intersection to obtain the color of the pixel for that ray. 

The sphere tracing process can be described as follows: given a ray, $\boldsymbol{r}$, originating at point, $\boldsymbol{p_0}$, iterative marching along the ray is performed to obtain its intersection with the surface. In the first iteration, this translates to taking a step along the ray with a step size of $\textit{uDF}(\boldsymbol{p_0})$ to obtain the next point $\boldsymbol{p_1} = \boldsymbol{p_0} + \boldsymbol{r}*uDF(\boldsymbol{p_0})$. Since $uDF(\boldsymbol{p_0})$ is the smallest distance to the surface, the line segment $[\boldsymbol{p_0}, \boldsymbol{p_1}]$ of the ray is guaranteed not to intersect the surface ($\boldsymbol{p_1}$ can touch but not transcend the surface). The above step is iterated $i$ times till $\boldsymbol{p_i}$ is $\epsilon$ close to the surface. The i-th iteration is given by $\boldsymbol{p_i} = \boldsymbol{p_{i-1}} + uDF(\boldsymbol{p_i})$ and the stopping criteria  $uDF(\boldsymbol{p_i})\leq\epsilon$. 

\begin{figure}
    \centering
    \includegraphics[width=\columnwidth]{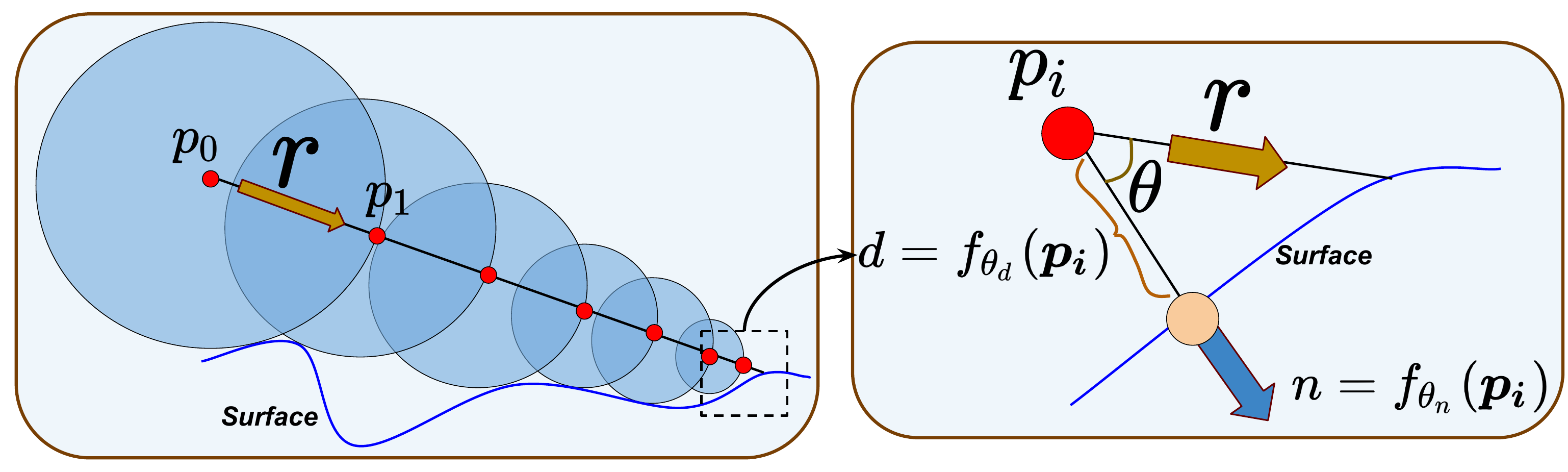}
    \caption{\textit{\textbf{Left:}} An illustration of the Sphere Tracing procedure described in Section.~\ref{sec:spheretracing}. \textit{\textbf{Right:}} Leveraging the \textit{nVF} for obtaining more accurate ray-scene intersections.}
    \label{fig:sphere_trace}
\end{figure}

Note that for \textit{uDF}s, the above procedure can be used to get close to the surface but doesn't obtain a point on the surface. One we are close enough to the surface, we can use a local planarity assumption (without loss of generalization) to obtain the intersection estimate. This is illustrated in Figure~\ref{fig:sphere_trace} and is obtained in the following manner:  if we stop the sphere tracing of the \textit{uDF} at a point $\boldsymbol{p_i}$, we evaluate the \textit{nVF} as $\boldsymbol{n} = nVF(\boldsymbol{p_i})$, and compute the cosine of the angle between the \textit{nVF} and the ray direction. The estimate is then obtained as $\boldsymbol{p_{proj}} = \boldsymbol{p_i} + \boldsymbol{r}*uDF(\boldsymbol{p_i})/(\boldsymbol{r}\cdot\boldsymbol{n})$. We also implemented other variants of the sphere tracing algorithm as detailed in (Section.~\ref{sec:exp_sphere_trace}) and found this approach to be the computationally efficient method and providing the best accuracy.



\subsection{From uDFs to Meshes.}
Sphere tracing \textit{uDFs}, described in the previous section, can only be used to render a view of the shape. Thus, the extracted iso-surface is immutable and cannot be used for applications such as 3D shape modeling, analysis and modification ~\cite{laga20183d}. Explicit 3D mesh representations are more amenable to such applications. In this section, we propose an approach to extract a 3D mesh out of the learnt \textit{uDF}.

A straightforward way to extract a mesh from a \textit{uDF} is to create a high-resolution 3D distance grid, followed by marching cubes on this grid. However, as discussed in~\cite{occupancy_network} this process is computationally expensive at high-resolutions, as we need to densely evaluate the grid. In~\cite{occupancy_network} a method for multi-resolution iso-surface extraction technique is proposed by hierarchically creating a binary occupancy grid by conducting inside/outside tests for a binary classifier based implicit representation. 

\begin{figure}
    \centering
    \includegraphics[width=\columnwidth]{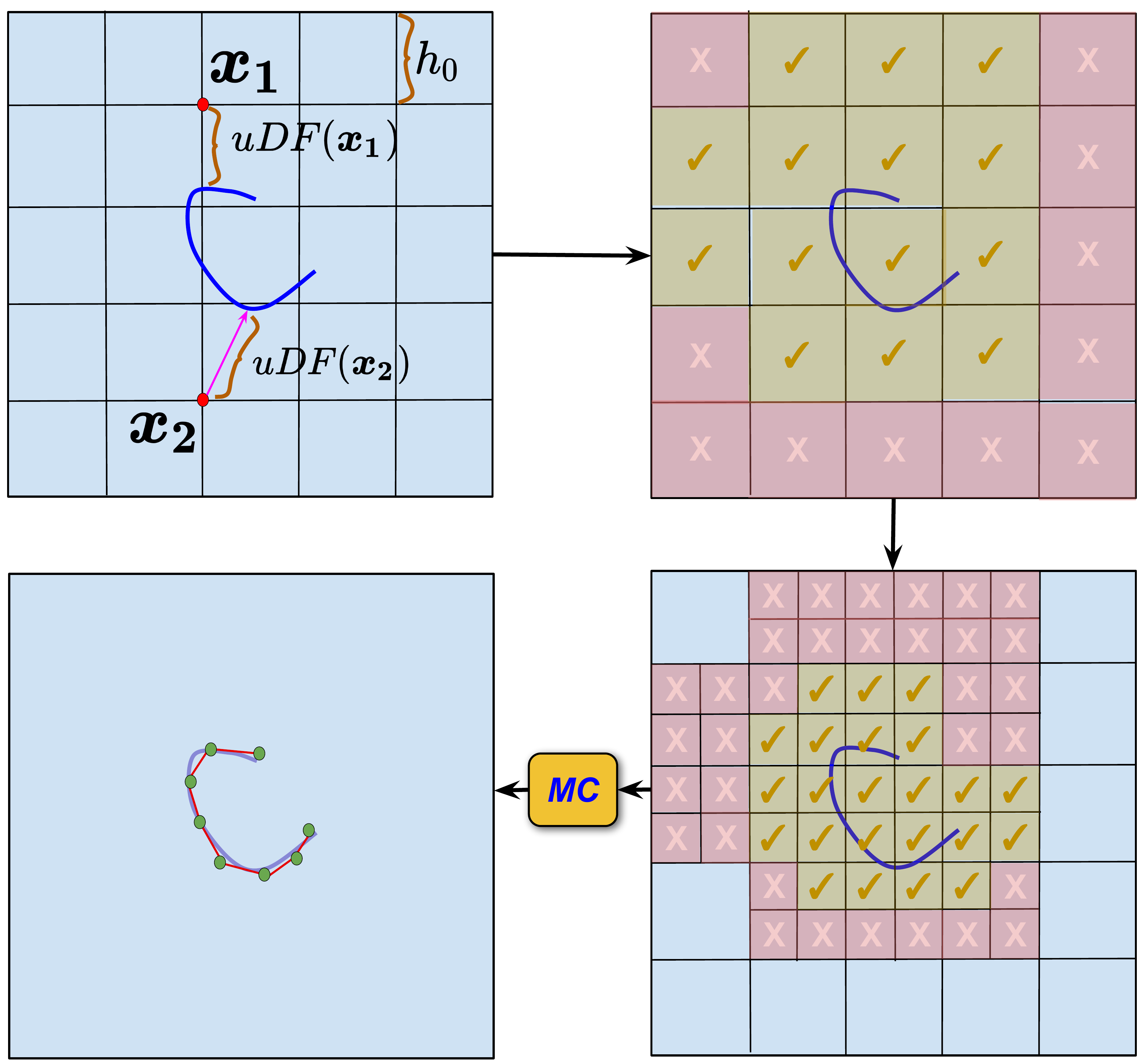}
    \caption{Multi-Resolution Isosurface extraction for \textit{uDFs} (described in Section.~\ref{sec:MISE}). At each level of heirarchy, we show the voxels selected for subdivision. Note that as we are using a \textit{uDF} and not an \textit{sDF}, there are a some false positive voxels selected close to the shape. Notice that these get eliminated in the next hierarchical level (Step 2 to Step 3).We perform Marching Cubes (MC) at the final resolution.}
    \label{fig:marching_cubes}
\end{figure}

However, in our case we only have a \textit{uDF} and cannot perform inside/outside tests. Therefore we propose a novel technique to hierarchically divide the distance grid using edge lengths of the cube. We illustrate the procedure in Fig.~\ref{fig:marching_cubes}. Starting with a voxel grid at some initial resolution, we first obtain a high resolution distance grid, then, perform marching cubes on that unsigned distance grid using a small positive threshold to get the final mesh. 
A voxel is chosen for subdivision if any of its eight corners have the predicted \textit{uDF} value $f_{\theta_d}(x) < h_i$, where $h_i$ is the edge length of the voxel grid at the $i'th$ level. The voxels that are not chosen for subdivision are simply discarded in the next level. Using this procedure, we quickly obtain a high-resolution distance grid, which is converted to a mesh using marching cubes.
In Fig.~\ref{fig:marching_cubes} we also describe some of the false positive subdivisions that our method generates and how we overcome them.

\section{Experiments}
\label{sec:experiments}
To reiterate, we hypothesized that the proposed method of learning shape representation is theoretically capable of representing complex arbitrary shapes with high-fidelity. In this section, we demonstrate the validity of these claims empirically.  We first select a set of prior work to compare against our method. Following this, we describe the evaluation metrics we use in our experiments. Finally, we present qualitative and quantitative results on learning implicit shape representations using \textit{DUDE}. 

\noindent
\textbf{Selection of prior art.} We pick two recent methods for learning implicit representations for shapes. First, we choose DeepSDF~\cite{deepsdf}, which is dependent on a watertighting process for learning shape representation which results in loss of fidelity. Second, we choose SAL~\cite{sal-2020}, which can learn implicit representations directly from triangle soups.  We demonstrate the limitation of SAL in representing  open shapes. In the following sections we present how our method overcomes the limitations of SAL and DeepSDF.

\subsection{Evaluation metrics}
\label{sec:eval metrics}
Before discussing the quantitative comparisons, we first introduce the metrics we use for evaluation. 

\noindent
\textbf{Depth Error.} First we evaluate the \textit{MAE} between the ground truth depth map and the estimated depth map which is obtained by Sphere Tracing the learnt representation. This error is evaluated only on the ``valid" pixels, which we define to be the pixels which have non-infinite depth in both the ground truth and estimated depth map. This metric captures the \textit{accuracy of ray-isosurface intersection}.

\noindent
\textbf{Normal Map Error.} Similarly, we evaluate the L2 distance between the sphere-traced normal map and the ground truth normal map for the valid pixels. Since the surface normals play a vital role in rendering, this metric is informative of the \textit{fidelity} of the final render.
    
\noindent
\textbf{IOU.} Finally, since both Depth Error and Normal map error are evaluated only on the valid pixels, they do not quantify whether the \textit{geometry of the final shape} is correct. Therefore, we also evaluate Pixel IOU, which is defined as,
    \begin{equation}
        IOU = \frac{\#\textit{Valid}\, \textit{Pixels}}{\#\textit{Invalid}\, \textit{Pixels} + \#\textit{Valid}\, \textit{Pixels}}
    \end{equation}
Here the Invalid pixels are those which have non-infinite depth in either the ground truth depth map or the estimated depth map but not both.

\subsection{Single Shape \textit{DUDE}}
In this section, we aim to empirically demonstrate the representational power of our model. To perform a comparative study, we train both SAL~\cite{sal-2020} and DeepSDF~\cite{deepsdf}. \\  

\noindent
\textbf{Choosing three challenging shapes.} To validate the benefit of our method, we evaluate on three challenging shapes. First we choose the \textit{Bathtub (B)}, which has high fidelity details. Second, we select the \textit{Split Sphere (S)}, to analyse how well our representation can model the gap between the spheres. Finally, we evaluate on \textit{Intersecting Planes (IP)}, a triangle soup with complex topology. Refer Fig.~\ref{fig:concept} for a visualization of these shapes.\\

\noindent
\textbf{Data Creation and training details.} We start off with a triangle soup normalized between [-0.5, 0.5], and densely sample 250000 points on the faces of the triangles. For each of these points, the associated normal for training the \textit{nVF} is the normal of the face from which it is sampled from. We randomly perturb these 250000 points along the xyz axes using the same strategy floowed in DeepSDF~\cite{deepsdf}. Now, for each of these 250000 points, we find the nearest point in the unperturbed set of points, and compute the distance between them. These distances are used to train the \textit{uDF}. Additionally we also sample 25000 points uniformly in the space, and follow the same procedure for creating the ground truth. Finally, we use 90\% of this data for training and 10\% for validation and train \textit{DUDE} using these samples. For both $f_{\theta_d}$ and $f_{\theta_n}$, we use a 6 layer MLP with ReLU activations and 512 hidden units in each layer. Adam optimizer is used with 1e-4 learning rate. Note, for training DeepSDF, we first convert the mesh to watertight using the same procedure followed in ~\cite{stutz2018learning} before sampling points in the space. 

\begin{table}[h!]
\centering
\resizebox{\columnwidth}{!}{
\begin{tabular}{c|c|c|c|c|c|c|c|c}
\hline
    \multirow{2}{*}{Shape}& \multirow{2}{*}{Method}& \multicolumn{3}{c|}{Depth Error} & \multicolumn{3}{c|}{Normal Error} & \multirow{2}{*}{IOU} \\
    \cline{3-8}
    & & Std. & Samp. & Proj.  & Std. & Samp. & Proj. &  \\
    \hline
    \multirow{3}{*}{B} & SDF~\cite{deepsdf} & 0.035 & 0.034 & 0.036 & 0.053  & 0.051 & 0.052 & 0.91 \\
    & SAL~\cite{sal-2020} & 0.028 & 0.027 & 0.029 & 0.065  & 0.067 & 0.069 & 0.92 \\
    & \textit{Ours} & \textbf{0.0154} & \textbf{0.0068} & \textbf{0.0066} & \textbf{0.028 } & \textbf{0.025} & \textbf{0.024 }& \textbf{0.96 }\\
    \hline
    
    \multirow{3}{*}{S} & SDF~\cite{deepsdf} & 0.0166 & 0.0161 & 0.0159 & 0.038  & 0.039 & 0.035 & 0.95 \\
    & SAL~\cite{sal-2020} & 0.0181 & 0.0173 & 0.0192 & 0.042  & 0.043 & 0.041 & 0.93 \\
    & \textit{Ours} & \textbf{0.0112} & \textbf{0.0054} & \textbf{0.0051} & \textbf{0.021 } & \textbf{0.019} & \textbf{0.019} & \textbf{0.98} \\
    \hline
    
    \multirow{3}{*}{IP} & SDF~\cite{deepsdf} & 0.023 & 0.019 & 0.021 & 0.065  & 0.061 & 0.063 & 0.94 \\
    & SAL~\cite{sal-2020} & 0.028 & 0.021 & 0.029 & 0.061  & 0.063 & 0.060 & 0.92 \\
    & \textit{Ours} & \textbf{0.0162} & \textbf{0.0072} & \textbf{0.0068} & \textbf{0.032}  & \textbf{0.029} & \textbf{0.023} & \textbf{0.97} \\
    \hline

\end{tabular}
}
\vspace{0.5mm}
\caption{Quantitative Comparison between DeepSDF~\cite{deepsdf}, SAL~\cite{sal-2020} and \textit{DUDE} on Depth Error, Normal Error and Silhouette IOU metrics described in Section.~\ref{sec:eval metrics}. Note we evaluate each method using the three Sphere Tracing strategies (Standard, Resample and Projection) discussed in Section.~\ref{sec:exp_sphere_trace}. B: Bathtub, S: Split Sphere, IP: Intersecting Planes}

\label{table:quant}
\end{table}

\noindent
\textbf{Quantitative Results.} We report the results on the different evaluation metrics introduced in Section.~\ref{sec:eval metrics} in Table~\ref{table:quant}. Note that SDF~\cite{deepsdf} is trained on the watertight mesh, but we evaluate the metrics on the original (non-watertight) mesh. For the bathtub mesh (B), we immediately find that the Depth Error of \textit{DUDE} is higher when compared to the other two methods. Similarly, for the Split Sphere (S), we see that SAL performs the worst, as it has a tendency to fill up gaps. Additionally, on the Intersecting Planes (IP), we again SAL to be marginally worse than DeepSDF. \\ 

\begin{figure}
    \centering
    \includegraphics[width=\columnwidth]{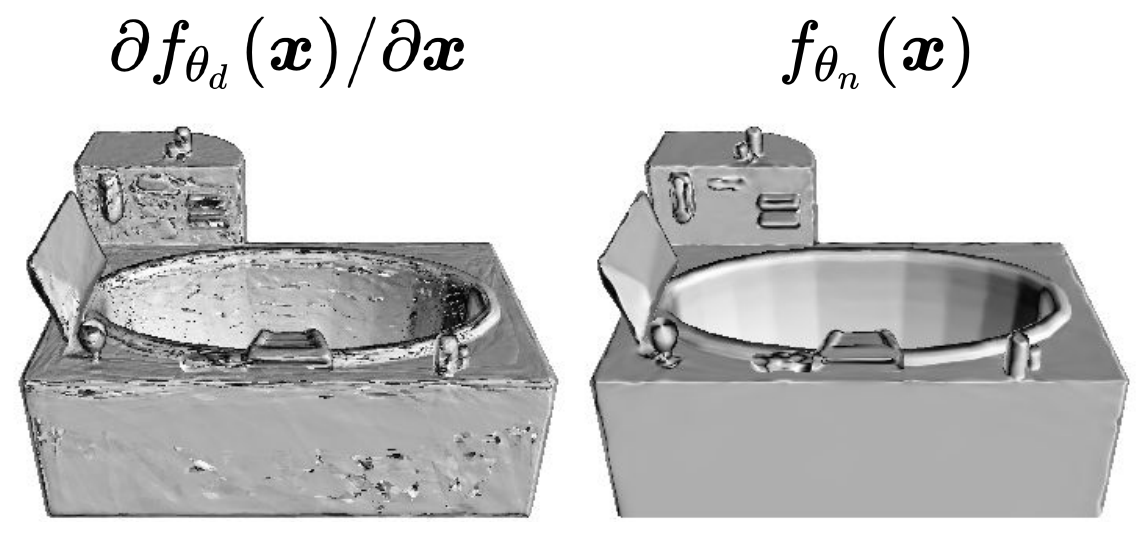}
    \caption{\textit{\textbf{Left:}} Ray casting using normals obtained by differentiating the \textit{uDF}. \textit{\textbf{Right:}} Ray casting using the surface normals estimated by the \textit{nVF}. This high-quality rendering empirically validates the need for learning an \textit{nVF} alongside a \textit{uDF}.}
    \label{fig:diff_nvf}
\end{figure}

\noindent
\textbf{Qualitative Results.} We present these results in Fig.~\ref{fig:concept}. These results are obtained by Sphere Tracing the learnt representation. For \textit{DUDE}, we use the \textit{nVF} to obtain the surface normals, and for the other methods, we differentiate the learnt distance field. We note some interesting findings. For the Bathtub model, a loss in fidelity is observed due to the watertighting process. SAL seems to retains some of the details, but the normal estimation fails, leading to low quality rendering. For the Split Sphere, no loss is fidelity is recorded owing to the simplicity of the shape, but the watertighting process distorts the geometry to cause inconsistent surface normals. This is primarily the reason why the renderings from DeepSDF don't look visually pleasing. On the other hand, SAL appears to close out the split sphere. However, for \textit{DUDE}, we obtain visually pleasing results, owing to high quality surface normals obatined from the learnt \textit{nVF}. A similar trend follows for the Intersecting Planes model. It is clearly evident that merely differentiating the learnt distance fields near the surface does not leads to imprecise surface normals estimates (See Fig.~\ref{fig:diff_nvf}), validating the need for learning a separate \textit{nVF}.

\begin{figure}[t]
    \centering
    \includegraphics[width=\columnwidth]{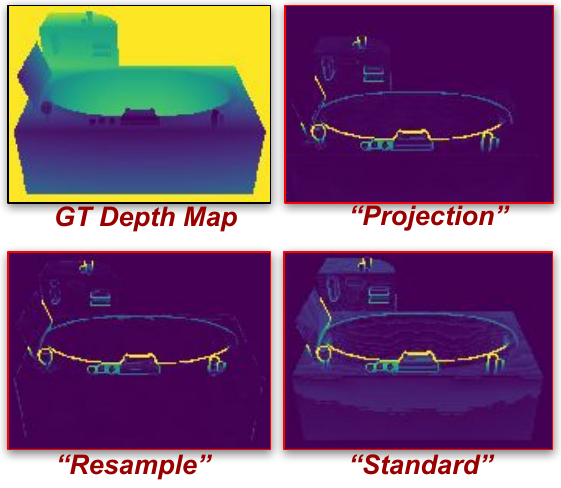}
    \caption{Absolute Depth Error between Ground Truth Depth Map and Depth Map estimated using different Sphere Tracing strategies described in Section.~\ref{sec:exp_sphere_trace}. Note that the \textit{``Resample"} strategy uses 100x more compute than the proposed \textit{``Projection"} startegy, but still has marginally higher error.}
    \label{fig:depth_error}
\end{figure}

\subsubsection{Sphere Tracing strategies}
\label{sec:exp_sphere_trace}
In this section, we describe baselines for the sphere tracing process, and demonstrate the superiority of the proposed approach discussed in Section.~\ref{sec:spheretracing}.

\noindent
\textbf{Baselines.} We consider two baselines for sphere tracing. First we consider a \textit{``Standard"} method, where we terminate the Sphere Tracing process on reaching a certain threshold. Second, after stopping the sphere tracing at a certain threshold, we resample the learnt uDF at 100 points along the direction of the ray in the vicinity of the point where we stopped the sphere tracing. More concretely, we stop the tracing process at $\boldsymbol{p_i}=\boldsymbol{p_{i-1}}+\boldsymbol{r}*f_{\theta_d}(\boldsymbol{p_{i-1}})$, and select a set of points $\mathcal{P} =\{p_i+\lambda \boldsymbol{r}\}$, by choosing 100 values of  $\lambda's$ uniformly in the range $[-0.01, +0.01]$. The point of intersection is given by $\underset{p \in \mathcal{P}}{\operatorname{argmin}}f_{\theta_d}(p)$. We call this second method as \textit{``Resample"}. Note that it takes 100x more time than the standard method. We call our proposed strategy as \textit{``Projection"}, which is described in Section.~\ref{sec:spheretracing}. 

\noindent
\textbf{Results.} In Table~\ref{table:quant}, we can clearly see that the \textit{``Projection"} method outperforms the \textit{``Standard"} method, and performs marginally better than the \textit{``Resample"} method which takes 100x more computation. In some cases, such as for (IP), the normal estimates are also improved as a result of the improved intersection computation. We also observe in the depth error maps shown in (Fig.~\ref{fig:depth_error}) that the \textit{``Projection"} method performs better qualitatively as well.

\begin{figure}[hb]
    \centering
    \includegraphics[width=\columnwidth]{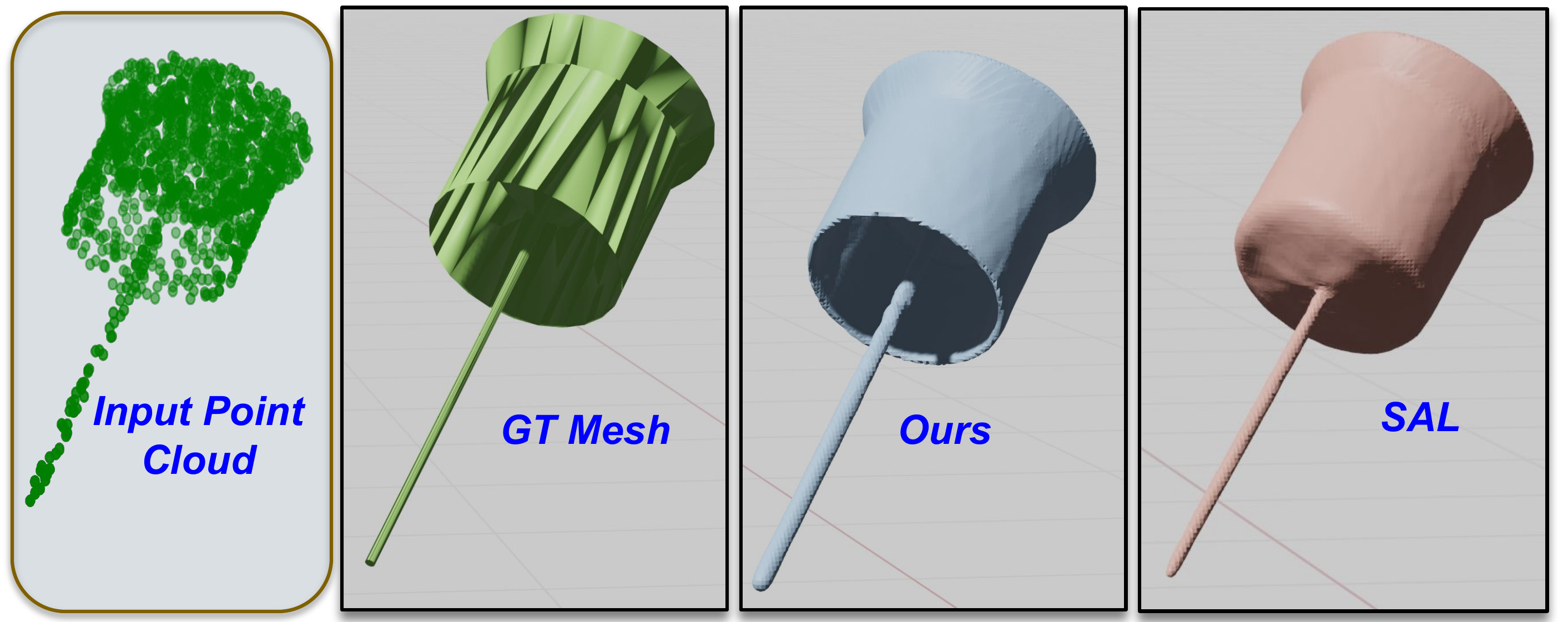}
    \caption{Results on surface reconstruction from point clouds for SAL~\cite{sal-2020} and \textit{DUDE}. Notice how \textit{DUDE} correctly models the gap in the input point cloud, while SAL tries to close up the gap. Note that the meshes shown in this figure for \textit{DUDE} are obtained by performing the iso-surface extraction process described in Section.~\ref{sec:MISE}. }
    \label{fig:lamp}
\end{figure}

\section{Learning a Shape Space}
\label{sec:shape space}
As discussed in prior work~\cite{deepsdf, sal-2020, occupancy_network}, learning a latent shape space for implicit functions is quite useful in practice, as it allows us to directly obtain a representation for unknown/unseen shapes without additional training. In this work, to demonstrate generalizability, we explore the problem of surface reconstruction from point clouds. More formally, we are interested in learning functions,
\begin{equation}
    \begin{gathered}
    f_{\theta_d}(\boldsymbol{z_i}, \boldsymbol{x}) \approx uDF_i(\boldsymbol{x}), and \\
    f_{\theta_n}(\boldsymbol{z_i}, \boldsymbol{x}) \approx nVF_i(\boldsymbol{x}) \\
    \end{gathered}
\end{equation}

Here, $z_i$ is the encoding of the sparse point cloud of the shape. Once trained on a set of training point clouds, we can evaluate the functions on unseen point clouds, and reconstruct the surface. 

We consider a subset of models from the lamp class of ShapeNet~\cite{chang2015shapenet}, and train both \textit{DUDE} and SAL. We show qualitative results in Fig.~\ref{fig:lamp}. Notice how SAL closes out the bottom of the lamp, whereas our model correctly models the gap. Additionally, on a held out test set, we obtain a mean chamfer distance of \textit{1.67e-3} as opposed to SAL which obtains \textit{1.84e-3}. This clearly demonstrates the superiority of our method in representing open shapes.

\section{Conclusion}
We demonstrated that our proposed shape representation, \textit{DUDE}, can effectively represent both open and closed shapes, with arbitrary and complex topology, with high fidelity. We showed, using both qualitative and quantitative evaluations, that our representation overcomes the failures of previous work such as SAL~\cite{sal-2020} which learns to close out shapes which are open, and methods such as DeepSDF~\cite{deepsdf} whose fidelity is limited by the watertighing process used to learn the representation. The presented work can to utilized to develop novel approaches that seek to model and exploit such complex 3D shapes. Possible avenues for future work include learning class agnostic latent shape spaces, and, learning representations with 2D supervision instead of triangle soups.


{\small
\bibliographystyle{ieee_fullname}
\bibliography{egbib}
}
\end{document}